\documentclass[10pt,twocolumn,letterpaper]{article}
\usepackage{cvpr}
\usepackage{times}
\usepackage{epsfig}
\usepackage{graphicx}
\usepackage{amsmath}
\usepackage{amssymb}
\usepackage{MYdefs}
\usepackage{gendefs}
\usepackage[pagebackref=true,breaklinks=true,letterpaper=true,colorlinks,bookmarks=false]{hyperref}

\cvprfinalcopy 

\ifcvprfinal\fi
\begin{document}
%

\title{Coupling Top-down and Bottom-up Methods for 3D Human Pose and Shape Estimation from Monocular Image Sequences}

\author{Atul Kanaujia\\
ObjectVideo, Inc.\\
Reston, VA\\
{\small atul.kanaujia@gmail.com}
}

\maketitle
\thispagestyle{empty}

\begin{abstract}
Until recently Intelligence, Surveillance, and Reconnaissance (ISR) focused on acquiring behavioral information of the targets and their activities. Continuous evolution of intelligence being gathered of the human centric activities has put increased focus on the humans, especially inferring their innate characteristics - size, shapes and physiology. These biosignatures extracted from the surveillance sensors can be used to deduce age, ethnicity, gender and actions, and further characterize human actions in unseen scenarios. However, recovery of pose and shape of humans in such monocular videos is inherently an ill-posed problem, marked by frequent depth and view based ambiguities due to self-occlusion, foreshortening and misalignment. The likelihood function often yields a highly multimodal posterior that is difficult to propagate even using the most advanced particle filtering(PF) algorithms. Motivated by the recent success of the discriminative approaches to efficiently predict 3D poses directly from the 2D images, we present several principled approaches to integrate predictive cues using learned regression models to sustain multimodality of the posterior during tracking. Additionally, these learned priors can be actively adapted to the test data using a likelihood based feedback mechanism. Estimated 3D poses are then used to fit 3D human shape model to each frame independently for inferring anthropometric biosignatures. The proposed system is fully automated, robust to noisy test data and has ability to swiftly recover from tracking failures even after confronting with significant errors. We evaluate the system on a large number of monocular human motion sequences.
\end{abstract}
\section{Introduction}
\begin{figure}
\begin{center}
\includegraphics[width=0.98\linewidth]{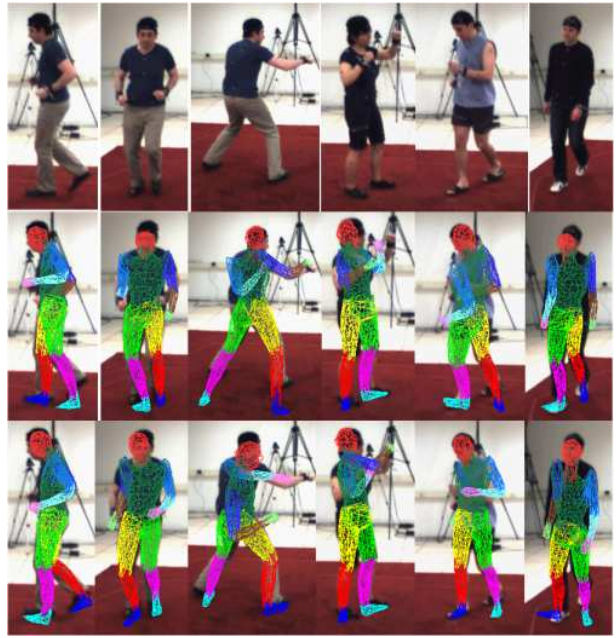}
\end{center}
\caption{Ambiguous observations with one-to-many correspondences between the 2D image and the 3D pose. The likelihood function based on silhouette overlap gives similar likelihood scores for these images}
\label{fig:multivalued}
\end{figure}

\begin{figure*}
\begin{center}
\includegraphics[width=0.98\linewidth]{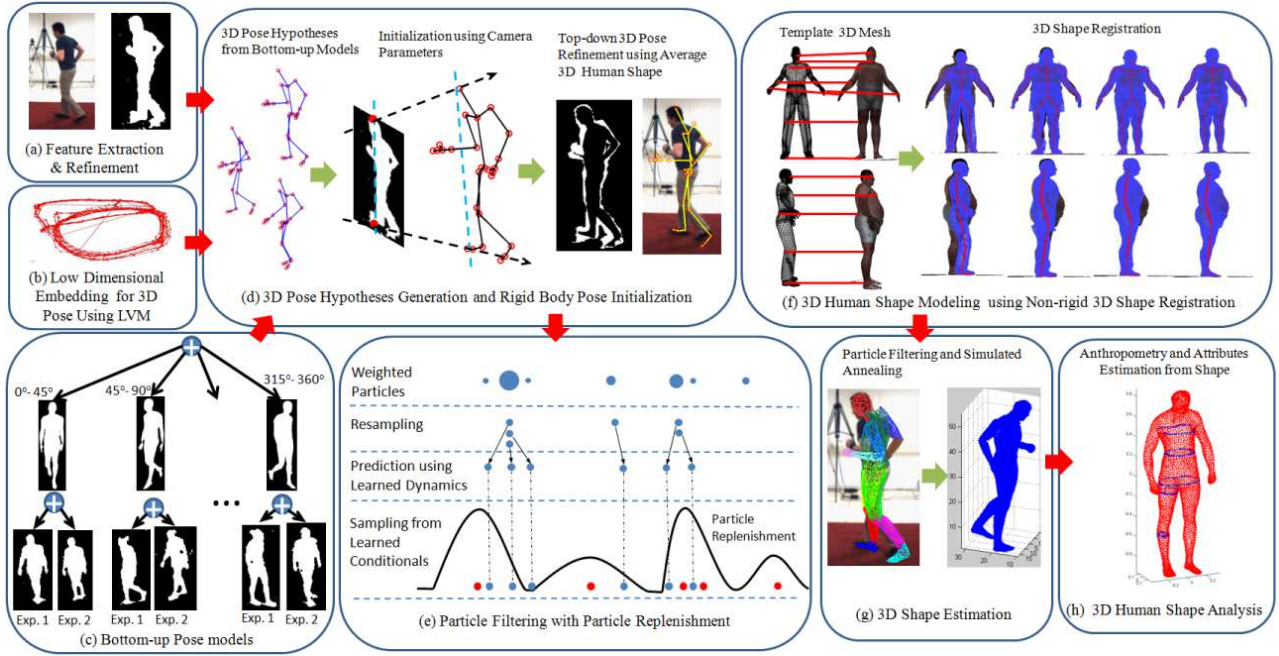}
\end{center}
\caption{Overview of our system for 3D human pose and shape estimation and analysis from monocular image sequences}
\label{fig:overview}
\end{figure*}

Extracting biosignatures from fieldable surveillance sensors is a desired capability for human intelligence gathering, identifying and engaging in human threats from a significant standoff distances. This entails fully automated 3D human pose and shape analysis of the human targets in videos, recognizing their activities and characterizing their behavior. However 3D human pose and shape inference in monocular videos is an extremely difficult problem, involving high dimensional state spaces, one-to-many correspondences between the visual observations and the pose states, strong non-linearities in the human motion, and lack of discriminative image descriptors that can generalize across a hugely varying appearance space of humans. Traditionally, top-down {\it Generative modeling} methods had been employed to infer these high-dimensional states by generating hypotheses in anthropometrically constrained parameter space, that get continuously refined by image-based likelihood function. However top-down modeling, being a somewhat indirect way of approaching the problem, faces challenges due to the
computationally demanding likelihood function and its requirement of accurate physical human models to simulate and differentiate ambiguous observations (see \fig{fig:multivalued}). Failure of top-down models have motivated the development of bottom-up, {\it Discriminative} methods - fast feed-forward approaches to directly predict states from the observations using learned mapping functions. Bottom-up methods, while being simple to apply, are frequently plagued by lack of representative features to model foreshortening, self and partial occlusion which limit their performance in unseen scenarios. In this work we attempt to combine the two approaches under a common framework of non-parametric density propagation system based on particle filtering. \Fig{fig:overview} shows the key components of 3D pose tracking and human shape analysis system. 

Particle filtering forms a popular class of Monte Carlo simulation methods for approximately and optimally estimating non-Gaussian posteriors in systems with non-linear measurements and analytically intractable state transfer functions. Intrinsically, particle filtering is a non-parametric generative density propagation algorithm, involving recursive prediction and correction steps to estimate the posterior over the high dimensional state space from a temporal sequence of observations. 
Generative simulation filters have been widely applied to various tracking problems in vision and are well understood. However, techniques to overcome their drawbacks, such as irrecoverable tracking failure due to noisy observations, by incorporating discriminative(predictive) cues, are less well explored. We develop three principled techniques to incorporate discriminative cues into the particle filter based 3D human pose tracking framework. The techniques are aimed towards overcoming limitations of particle filtering by improving both the proposal density modeling as well as the likelihood computation function. Unlike past approaches, we use bottom-up methods to not only initialize and provide discriminative cues to the top-down methods for improved tracking, but also have a feedback mechanism to adapt bottom-up models using online learning from top-down modeling.  
In Particle Filtering(PF), sampling from a marginal distribution is made tractable by recursively computing particle weights, causing degeneracy of particles. This is efficiently overcome using re-sampling, which however, over longer sequences, causes sample impoverishment problem. This is a more difficult problem and currently no principled mechanism exist to overcome it. In context of 3D human pose tracking, lack of particle diversity may cause failure to preserve multimodality of the posterior density. \Fig{fig:multivalued} illustrates the severity of tracking failures due to inability to track all the modes using a particle filtering framework. 
In this work we tackle the problem of characterizing the multivaluedness of the dataset and develop algorithms to sustain it during tracking. We identify the cause of the sample impoverishment as the underlying generative, model-based tracking mechanism of the PF, that cannot model large deviations from the examples that are typical to a data set. In contrast, predictive (discriminative) methods  provide an alternative approach to handle difficult cases not modeled by generative methods. Specifically, learned priors can be used to furnish additional particles during the tracking process and maintain multimodality for enhanced 3D pose recovery. Although pure discriminative methods such as \cite{SKLM05} had been used in the past to propagate the posteriors at each time step using a learned conditional, our work attempts combine the two approaches in a mutually complementary framework and overcoming setbacks in one using strengths of the other.     
The tracked 3D pose are then used to estimate 3D human shapes using simulated annealing. The 3D shapes, estimated independently at each frame, are then used for inferring attributes such gender, weight, height and dimensions of various body parts by averaging over the sequence of frames. In principle, the tracked 3D poses in a sequence of frames can be used to iteratively infer the posterior over 3D shape parameters using a forward-backward algorithm. We anticipate that this extension will improve shape analysis compared to current framework and is planned as future work during the course of project. We provide extensive evaluation results for various components of the system, and demonstrate the efficacy of our algorithms in overcoming strong ambiguities observed in the data. 
 
\noindent\textbf{Contributions:} (a) We develop a fully automated system for estimating and analyzing shapes of humans in monocular video sequences ; (b) We develop a novel measure to characterize multimodality in the dataset ; (c) We develop principled techniques to incorporate predictive cues in the particle filtering framework and preserve  multimodality for the 3D pose tracking;  (d) Demonstrate principled integration of online learning into tracking framework. The learning progressively adapts the predictive models to the dataset by including accurately predicted examples in the basis set and reducing errors due to training bias. 
 
\noindent\textbf{System Overview:} \Fig{fig:overview} sketches the control flow diagram and lists the key components of our system : (a) Silhouettes extracted using background subtraction are used to compute shape descriptors ; (b) Low-dimensional representation of 3D human body pose is learned offline using  non-linear Latent Variable Model(LVM) \cite{KSM07b}; (c) Bottom-up(discriminative) models are trained offline from the labeled examples obtained from the motion capture data. Training involves learning hierarchical mixture of experts by partitioning the data set based on viewpoint at level 1 and one-to-many mapping ambiguity at level 2; (d) Bottom-up models are used to initialize the global orientation and 3D joint angles (pose) from the 2D silhouette shape. Translation in 3D space is estimated using the camera calibration parameters. Joint angles and global orientation are optimized using simulated annealing and average 3D shape; (e) Tracking is performed using annealed particle filtering by sampling particles both from the dynamics and the bottom-up proposal distributions; (f) Statistical 3D human shape model is learned offline by non-rigidly deforming a 3D template mesh to laser scan data; (g) 3D Shape is fitted to the observed silhouettes using annealed particle filtering ; (f) Estimated 3D shape is used to extract biosignatures and physiological attributes of the human.

\section{Related Work}
Since the introduction about 20 years ago, particle filtering have been widely applied in various domains of target tracking and optimization problems. A comprehensive tutorial on various particle filtering methods is given in \cite{Doucet11}\cite{Doucet00}\cite{AMGC02}. A number of enhancements of particle filtering already exist in literature (such as Auxiliary PF, Gaussian Sum PF, Unscented PF, Swarm Intelligence based PF and Rao-Blackwellised Particle Filtering for DBN) specifically focused on setbacks of simulation based filtering. Although only a few of the works have addressed possible ways of incorporating discriminative information in the filtering process. Some of the relevant works that have attempted to combine the two approaches in the past include \cite{SKM06,SBB07} for articulated body pose recovery in static images, \cite{IB98a,DBR00,LC04} for improving tracking and \cite{SU10} for non-rigid deformable surface reconstruction. Sminchisescu {\it et} al.\cite{SKM06} proposed an efficient learning algorithm to combine the generative and the discriminative information by incorporating a feedback mechanism from the generative models to improve predictions of the discriminative model. Rosales and Sclaroff \cite{RS01} employed discriminative model based on mixture of neural networks as a verification step to the generative pose estimation in static images. 
Urtasun {\it et} al.\cite{SU10} proposed a combined framework of the two approaches by explicitly constraining the outputs of discriminative regression methods using additional constraints learned as a generative model. Notable among these are the approaches proposed in \cite{IB98a,DBR00,LC04,SBB07} that employ simulation based methods to recover 3D pose in monocular image sequences. Sigal {\it et} al.\cite{SBB07} used discriminative models as an initialization step for the pose optimization problem for static images. Lee and Nevatia \cite{LN09} developed 3D human pose tracking using data driven MCMC. They used a rich combination of bottom-up belief maps as the proposal distribution to sample pose candidates in their component-based Metropolis-Hastings approach. However their  mixing ratios are predetermined and chosen in ad hoc fashion. Whereas we propose a more principled approach to adaptively determine these ratios to overcome specific limitations of simulation based filters. The approach to combine top-down and bottom-up information in \cite{IB98a,DBR00} also employs a pre-defined importance sampler from data-driven belief map using distributions are not learned, a significantly different approach than ours. Sustaining multimodality in particle filtering domain has been addressed in the past by Vermaak {\it et} al.\cite{VDP03}, albeit in context of limiting the re-sampling to a set of mixture components fitted to the particles. The approach may be somewhat restrictive if number of mixture components are too small. To the best of our knowledge no principled mechanism to combine the discriminative and generative information to overcome deficiencies of the PF tracking method while simultaneously adapting the predictive priors have been proposed in past. Active online learning has been widely applied in all major learning based frameworks of computer vision. However, none of the works in the past have applied it in context of the problem of 3D human pose recovery from monocular image sequence. 
\section{Measure of Multimodality} 
\begin{figure}
\begin{center}
\includegraphics[width=0.98\linewidth]{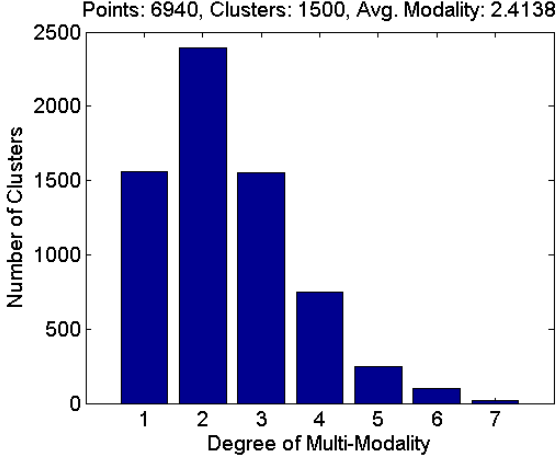}
\end{center}
\caption{Degree of Multimodality between input(image features) and output(3D joint angles) for the HumanEva motion capture data \cite{SHSV10}  with $N = 6940$ data pairs and $N_{clusters}= 1500$}
\label{fig:multivaluedPlot}
\end{figure}
We develop an information theoretic measure to quantify multivaluedness of mapping from 2D image to 3D pose in a human motion capture dataset. Our approach extends the multimodal data representation presented in \cite{SKLM05} which models the degree of multivaluedness present in the data as number of unique 3D pose ($\xx_t$) clusters in correspondence with the elements in the 2D image ($\rr_t$) clusters. The clusters obtained using K-Means for both $\rr_t$ and $\xx_t$ model perturbations due to noise in the observation and pose space respectively. Fig. \ref{fig:multivaluedPlot} shows the histogram obtained from the multimodality analysis of the data. We make two modifications to the weights given to these associations: (a) Large observation(input) cluster sizes associated to single or multiple pose(output) clusters implies stronger multimodality in the dataset. We explicitly give weights proportional to cluster sizes in generating the histogram; (b) Within each cluster, the distribution of points associated to different clusters reflects the multimodality of the data. For example, an input cluster with output cluster association indices as $A = [1,1,1,1,2,3]$ reflects weaker tri-modality compared to $B = [1,1,2,2,3,3]$ even when both have same cluster sizes. We encode this information using the Shannon's Entropy $H(x) = -\sum_i p(x_i) \mbox{log}_2 p(x_i) $ that captures the regularity of the probability distribution of the input cluster points to be associated to different output clusters. For our case  $ H(A) = 1.8136 $ and $H(B) = 2.1972 $. The weights for the correspondence between the input clusters ($\xx$) and the output clusters ($\yy$) can be formulated as :
\begin{equation} 
\label{eq:multimodality}
h_{n}(\xx) =  N(\xx) \frac{\mbox{exp}(-\sum^n_i p(\yy_i) \mbox{log}(p(\yy_i)))}{n} 
\end{equation}
where $h_n(\xx)$ is the weights attached to the associations between $N(\xx)$ elements of the $\xx$ cluster and  the corresponding outputs $\yy$ spread across $n$ clusters. Fig. \ref{fig:multivaluedPlot} shows the multimodality plot obtained from HumanEva dataset\cite{SHSV10} for $N=6940$ data points and $N_{clusters}= 1500$. As discussed in the experiment section \ref{sec:experiments}, we use this measure to quantifying the degree of multimodality maintained by the tracked hypotheses in the particle filtering. 
\section{Predictive Models for 3D Human Poses} 
\label{sec:discriminative}
We work with temporally ordered sequence of vectors $\cY_n = \{\yy_1,\cdots,\yy_n\}$ denoting 3D human body pose as a vector of joint angles, $\cX_n=\{\xx_1,\cdots,\xx_n\}$ as the states (latent space of the 3D pose vectors) learned using non-linear latent variable model and $\cR_n = \{\rr_1,\cdots,\rr_n\}$ as the image observations in the form of silhouettes obtained using background subtraction. We use Spectral Latent Variable Model(SLVM)\cite{KSM07b} to compute the low-dimensional representations of the pose vectors. In principle, any latent variable model (such as GPLVM, GPDM and GTM) that supports structure preserving, bi-directional mappings, can be used here for removing correlations between redundant dimensions of joint angle space. 
We work in the latent space of $\cX$ both for learning predictive models and filtering. The original joint angle space $\cY$ is used for likelihood computation, 3D pose visualization and rendering. 
To preserve diversity of the particles and multimodality of the posterior, we replenish the particles by sampling from a multimodal prior learned as hierarchical Bayesian Mixture of Experts (hBME) to model multivalued relation between 2D image space to 3D human pose space. In hBME, each expert(functional predictors) is paired with an observation dependent gate function that scores its competence in predicting states when presented with different inputs(images). For different inputs, different experts may be active and their rankings (relative probabilities) may change. The conditional distribution over predicted states has the form: 
\begin{equation}\label{eq:exp_model}\nonumber
p(\xx| \rr,\Om)=\sum^{N_v}_{v=1} g_v(\rr | \gm_v)\sum^{N_d}_{i=1}  g_i(\rr | v, \lm_i) p_i(\xx | \rr,\WW_i,\SSigma_i^{-1}) 
\end{equation}
\noindent where $\Om = \{\WW,\gm,\lm,\SSigma\}$ are the parameters of the classification ($g_v$ and $g_i$) and regression functions. At the highest level, gate functions $g_v$ partition the data into $N_v$ view-specific clusters to model view-based ambiguities. Within each cluster we further partition the data into $N_d$ predictive sets to model depth-based ambiguities. For each set, we train regression models using Relevance Vector Machine\cite{T01} with the predictive distribution for the experts $p_i$ learned as Gaussian functions \eqref{eq:experts} centered at the expert predictions (non-linear regressors with weights $\WW_i$).
\begin{equation}\label{eq:experts}
p_i(\xx|\rr,\WW_i,\SSigma_i^{-1}) = \cN(\WW_i \Phi(\rr),\sigma_D + \Phi(\rr)\SSigma_i\Phi(\rr)^T)
\end{equation}
\noindent where the predictive variance is sum of fixed noise term in training data $\sigma_D$ and input specific variance $\Phi(\rr)\Sigma_i\Phi(\rr)^T$ due to uncertainty in the weights $\WW_i$. The gate functions $g_v$ and $g_i$ are the input dependent linear classifiers modeled as softmax functions with weights $\lm_i$ and are normalized to sum to 1 for any given input $\rr$. $g_i(\rr) = \frac{\exp(\lm_i^\top\rr)}{\sum_k \exp(\lm_k^\top\rr)}$, where $\rr$ are the image descriptors, $\xx$ state outputs in the latent space in correspondence to 3D pose $\yy$ in original joint angle space. The gate function $g_v(\rr | \gm_v)$ is  trained to recognize the view $v$ using observation $\rr$. Within each view $v$, $g(\rr | v,\lm_i)$ outputs the confidence of an using an expert for predicting the state. 

\noindent\textbf{Bayesian Online Learning for hBME:} Performance of predictive models depends on the assumption that training examples are representative of the test data. 
We develop an online learning algorithm to dynamically adapt predictive models to the test data during the generative filtering process. Accurate 3D pose hypotheses generated by the PF are used to improve the accuracy and specialize the predictive priors to the test domain. This involves both updating the parameters as well as adaptively updating the bases set of the of the gates and experts of the hBME. We use Bayesian relevance criteria to add/delete elements from the bases of the learned models, that attempts to maximize the marginal likelihood(ML) of the observation with respect to the hyper-parameters of the model. The hyper-parameters are the parameters of hierarchical priors that control the sparsity of the models using Automatic Relevance Determination(ARD) mechanism \cite{T01}. A new labeled data $(\rr_i,\xx_i)$ is included into a bases set of an RVM classification (or regression) if its inclusion improves the ML of the model. The decomposition of the covariance matrix aid the computation of the change in ML due to an individual element :
\begin{equation} 
| \SSigma^{-1} | = | \SSigma^{-1}_{-i} |  - \frac{\SSigma^{-1}_{-i} \Phi(\rr_i) \Phi(\rr_i)^T \SSigma^{-1}_{-i}}{\alpha_i + \Phi(\rr_i)^T \SSigma^{-1}_{-i}\Phi(\rr_i)}
\end{equation}  
\noindent where $\SSigma^{-1}$ and $\SSigma^{-1}_{-i}$ are the covariance with and without the new data, and $\alpha_i$ is the hyperparameter denoting the uncertainty of the weights $\WW_i$ of the new basis element to be zero. The change in the ML due to added basis element  $\cL(\alpha) = \cL(\alpha_{-i}) + l(\alpha_i)$ can be independently analyzed using $l(\alpha_i)$ to make decision about its inclusion in the basis set. Inclusion of a new basis may result in redundancy due to presence of other elements which can be consequently re-evaluated to support inclusion (or deletion) of other elements in the bases set. The new bases set are used to re-estimate the parameters of the models. 
\begin{figure*}[t]
\begin{center}
\includegraphics[width= 0.98\linewidth]{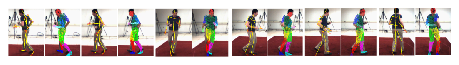} \\
\includegraphics[width=0.98\linewidth]{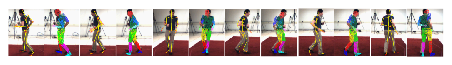} \\
\includegraphics[width=0.98\linewidth]{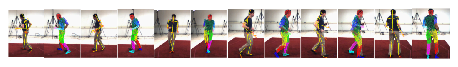} \\
\includegraphics[width=0.98\linewidth]{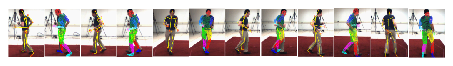}
\end{center}
\caption{Comparison of tracking results for the 4 different Particle Filtering algorithms on HumanEva I data set for Jogging sequence of Subject S2 for the frame 13, 34, 127, 174, 198 and 243. We show the estimated pose and the deformed average 3D human mesh model. All the particles were initialized using pose estimates from the bottom-up model in frame 13.{\it(Top row)} Tracking results using Annealed Particle Filtering(APF). Notice tracking failure in frame 34 and 198 due to left-right leg ambiguity and viewpoint ambiguity in frame 127 ; {\it (Second row)} Tracking results using {\it Optimal Proposal Density}. The learned distribution is accurately able to resolve ambiguities and prevents tracking failure ;{\it (Third row)} Tracking results using {\it Joint Particle Filtering} achieves best level of accuracy with accurate parts alignment with observation,{\it (Fourth row)} Tracking results using {\it Joint Likelihood Modeling }}
\label{fig:CompareAPF}
\end{figure*}

\section{Incorporating Predictive Cues in Annealed Particle Filtering}
Tracking is initialized using predictive models. Approximate translation is estimated using geometry, assuming no camera tilt and the human to be of height $1.78m$ in upright poses, from the calibration parameters of the camera. Generative filtering algorithm involves recursive propagation of the posterior over the state sequences at each time step $n$ using the following prediction and correction step $p(\cX_n | \cR_n) \varpropto$
\begin{equation}\nonumber
 p(\rr_n | \xx_n) \int p(\xx_n | \cX_{n-1},\cR_{n-1})p(\cX_{n-1} | \cR_{n-1}) d\xx   
\end{equation}
Particle filtering propagates the posterior as a set of $N_s$ weighted particles (hypotheses) at each time step $n$ as $\{\xx^i_n,\ww^i_n\}_{i=1\cdots N_s}$.
Particle Filtering computes these importance weights at successive time steps,  recursively using the weights in the previous time step. This avoids increasing computational complexity for recomputing weights for the entire sequence $\cX_n$ at every time step $n$: 
\begin{equation}
\ww^i_n = \ww^i_{n-1} \frac{p(\rr_n | \xx^i_n) p(\xx^i_n | \xx^i_{n-1})}{q(\xx^i_n | \cX^i_{n-1}, \cR^i_n)}
\end{equation} 
\noindent where the importance density at time step $n$ is further approximated as $q(\xx_n |\cX_{n-1},\cR_n) \approx q(\xx_n |\xx_{n-1},\rr_n)$  
Simulated Annealing(SA) is a stochastic optimization algorithm that runs a series of re-sampling and diffusion steps to attain an approximate global optima. APF, introduced by Deutscher et. al\cite{DBR00}, employs Simulated Annealing optimization at each time step to diffuse the particles to other modes of the cost function. 
We extend Annealed Particle Filtering(APF) algorithm to integrate predictive cues from the learned priors. APF approximates the importance density as $q(\xx_n | \xx^i_{n-1},\rr_n) = p(\xx_n | \xx^i_{n-1})$. The weight update equation thus becomes  $\ww^i_n \varpropto \ww^i_{n-1} p(\rr_n | \xx^i_n)$. The re-sampling and simulated annealing optimization are then performed alternately at each iteration.
\subsection{Optimal Proposal Filtering (OPF)}
The optimal importance density\cite{Doucet00} is given by $q_{opt}(\xx_n | \xx_{n-1},\rr_n) = p(\xx_n | \xx_{n-1},\rr_n)$. 
This density is called optimal as sampling it gives the following recursive update equation of the weights of the $i^{th}$ particle as $\ww^i_n \varpropto \ww^i_{n-1} p(\rr^i_n | \xx^i_{n-1})$ thus making the new weights effectively independent of the sampled particles $\xx^i_n$. Our first method for incorporating predictive information learns this distribution as conditional Bayesian Mixture of $M$ Experts (cBME). The form of the Bayesian Mixture of Expert model is similar to as discussed in Section \ref{sec:discriminative} with only one level of gate functions. A key issue in learning this conditional is to accurately model the relative scales of the state space data points $\xx_{t-1}$ and the observations $\rr_t$. This is required to avoid the prediction in the current frame to be entirely driven by either the current observation or the previous state. Therefore, for training the experts and gates in our BME, we use kernel basis of the form:
\begin{equation}
\Phi(\xx,\rr) = K_{\sigma_{\xx}}(\xx,\xx_i) K_{\sigma_{\rr}}(\rr,\rr_i) 
\end{equation}
where the rbf kernel has the form $K_{\sigma_{\xx}}(\xx,\xx_i) = e^{-\sigma_{\xx} \|\xx - \xx_i\|^2}$. The scales $\sigma_{\xx}$ and $\sigma_{\rr}$ determine how well the learned conditional $p(\xx_n | \xx_{n-1},\rr_n)$ is able to generalize to test examples. Too narrow width for $\xx$ may turn-off the kernels if the estimated pose from previous time step differs even slightly from the training exemplars, and may cause the predictors to output an average pose. If the scale is too wide, the regression model may simply average from the multiple observation based predictions. As is true in any predictive modeling, this method assumes that both train and test exemplars are representative samples from a common underlying distribution.
\subsection{Joint Particle Filtering (JPF)}
Importance density should be as close to the posterior to achieve optimal tracking performance. Already there exist techniques (based on partitioned sampling and bridging densities) to overcome sub-optimal proposal densities. Choosing an appropriate importance density can reduce the effect of sample impoverishment in particle filtering and consequently its ability to recover from errors. Narrower predictive distributions from the learned bottom-up models provide a useful proposal to generate particles with higher weights that can competently span the posterior state space. The predictive proposal distribution is a mixture of Gaussian summed across all the viewpoints and expert models represent all plausible poses for a given observation. At each time step we replace a few particles by the the samples from the predictive proposal $p_B(\xx_n | \rr_n)$ to maintain particle diversity.  The importance density is modeled as: 
\begin{equation}
q(\xx_n | \xx_{n-1},\rr_n) =  (1-\gamma_n)p(\xx_n | \xx_{n-1}) +  \gamma_n p_B(\xx_n | \rr_n)  
\end{equation}
Critical to this approach is to dynamically adjust the fraction $\gamma_n$ at each time step $n$. $\gamma_n$ acts as a balance between the predictively and dynamically sampled particles. This will enable effective recovery from errors during failure when the proposal density fails to generate any particles near the true posterior. In our experiments, we found the tracking accuracies to be greatly influenced by the fraction $\gamma_n$. A possible approach to estimate $\gamma_n$ is to use traditional data fusion model to combine probabilistic densities using Central Limit Theorem(CLT) that assigns the weights as inverse of their variances to the individual densities. Both the proposal densities from the learned dynamic model and the bottom-up models are probabilistic non-linear regression functions learned using Relevance Vector Machine(RVM)\cite{T01}. The proposal densities has the same form as specified in the eqn. (\ref{eq:experts}). The predictive variance for a test input $\rr$ is $\sigma = \sigma_D + \Phi(\rr)\Om\Phi(\rr)^T$. Here $\sigma_D$ is the fixed maximum likelihood estimate of variance due to the training data. The second data dependent term denotes the confidence of the regression function in the prediction from a given input $\rr$. CLT sets the fraction as $\gamma_n = \frac{\sigma_2}{(\sigma_1 + \sigma_2)}$ where $\sigma_1$  and $\sigma_2$ are the predictive variance of the dynamical model $p(\xx_n | \xx_{n-1})$ and the predictive proposal $p(\xx_n | \rr_n)$ respectively. In practice, however we found this approach less effective in balancing the particles to be sampled from either of the distributions. The fraction tends to be strongly dependent on the learned models rather than the observations. In an ideal scenario $\gamma_n$ should increase when particles sampled from the dynamic model has lower weights and should be at lower value when the dynamic model is doing well. We adopt a simplistic approach to control the number of particles sampled from the dynamics model and the discriminative map by adjusting the fraction purely based on weights of the particles sampled from either of the distributions. At each time step we compute the gamma as 
\begin{equation}
\gamma_n = \frac{\sum_{i \epsilon \cN^{BU}_{n-1}} \ww^{(i)}_{n-1}}{\sum_{i \epsilon \cN^{BU}_{n-1}} \ww^{(i)}_{n-1} + \sum_{i \epsilon \cN^{DYN}_{n-1}} \ww^{(i)}_{n-1}} 
\end{equation}
\noindent where $\cN^{BU}_{n-1}$ and $\cN^{DYN}_{n-1}$ denote the set of particles sampled from the predictive proposal map and dynamic distribution respectively. The $\ww^{(i)}_{n-1}$ are the particle weights before resampling in the previous time step. 
The motivation behind this weighting scheme is to assign high weights to the proposal which is generating particles closer to the true posterior. We initialize the $\gamma_0$ to 0.5 in the first time frame and at each time step update the fraction to adaptively control samples diversity. 
In principle, the dependence of the fraction $\gamma_n$ on the particle weights can be extended to include longer history of particle weights from the previous $N > 1$ frames. However, we found the current implementation to be sufficient yet significantly improve the accuracies of the APF tracker.
\subsection{Joint Likelihood Modeling (JLM)}
Likelihood distribution computes the belief of particles in the light of current observations. Ambiguities in 2D to 3D mappings are primarily due to failure of the likelihood function to assign different weights to seemingly similar but different 3D poses. Likelihood computation can be enhanced by incorporating richer low-level features that discriminative yet can generalize to different test scenarios. We propose an effective method to improve the likelihood model treating the learned conditional $p_B(\xx|\rr)$ from bottom-up models as a prior distribution over the state$\xx$ space conditioned on the input $\rr$. The joint likelihood is modeled as: 
\begin{equation} 
 p_L(\rr_n | \xx_n,\cH(\rr_n)) \varpropto  p(\rr_n | \xx_n)^{1 - \beta} p_B(\xx_n| \cH(\rr_n))^{\beta} 
\end{equation}
\noindent where $\cH$ denotes the extracted descriptor of the observed silhouette. The fraction $\beta$ that gives different weights to the likelihood distribution and the predictive conditional is chosen by cross-validation and is fixed to 0.35 in all the experiments. In our case, $p(\rr_n | \xx_n)$ is modeled as complex non-linear transformation of projecting a synthetic 3D mesh based computer graphic model of human in the pose $\xx$ (see section \ref{fig:overview}) and compute the degree of overlap between the projected and the observed silhouettes. Whereas the bottom-up models employ shape information ($\cH(\rr_n)$) extracted from the outer contour of the silhouette (shape context followed by vector quantization) that are in some sense complementary to the silhouette overlap information used in $p(\rr_n | \xx_n)$. As $p_B(...)$ has an analytical form of mixture of Gaussians (see eqn. (\ref{eq:exp_model})), it can be readily evaluated for any of the particle $\xx^{(i)}_n$. The conditional prior simply reweighs the likelihood cost based on how close the hypothesized state of the particle  $\xx^{(i)}_n$ is to the discriminatively predicted state $\hat{\xx}$. In theory, a linear combination of the two distributions (with adjustable weights) may also be used to compute likelihood weights of the particles. In the next section we compare the results of the extensive evaluation we performed for the three filtering algorithms with the baseline annealed particle filtering based tracker.  

\begin{figure*}[t]
\begin{center}
\includegraphics[width= 0.98\linewidth]{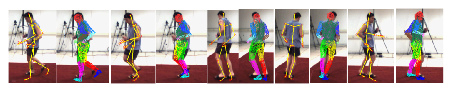} \\
\includegraphics[width= 0.98\linewidth]{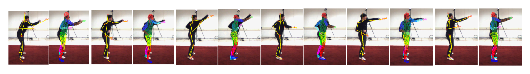} \\
\includegraphics[width=0.98\linewidth]{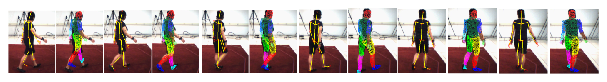} \\
\includegraphics[width=0.98\linewidth]{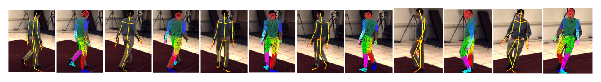} 
\end{center}
\caption{Tracking results using {\it Joint Particle Filtering} on sequences significantly different from the training data. {\it (First row)} HumanEva I sequence for the test subject S4 jogging in arbitrary orientation. JPF can successfully track the sequence as the learned dynamics does not include root orientation ; {\it (Second row)} Tracking results using {\it Joint Particle Filtering} on Boxing  and Walking {\it(Third row)}  sequence for the subjects S3 and S1 respectively ; {\it (Fourth row)} Tracking results on HumanEva II data set}
\label{fig:ResultsUnseenData}
\end{figure*}

\section{3D Human Shape Modeling and Estimation} 
Principal Component Analysis (PCA) is used to calculate global shape subspace that models variation of 3D human shapes of 1500 subjects. To learn the shape space, we register a template reference mesh model with 1200 vertices to CAESAR\cite{CAESAR} laser scan data to parameterize human body shapes as 3600 dimensional vector. This reference model is a hole-filled, mesh model with standard anthropometry and standing in the pose similar to the subjects in the CAESAR dataset. The CAESAR dataset has 73 landmark points on various positions, and these can be used to guide the 3D shape registration.  The registration process consists of the following steps: (1) Using the MAYA graphic software, we generate a reference mesh model that has a similar pose as the models in the CAESAR dataset ; (2) We annotate landmark points on this reference model (as illustrated in \ref{fig:overview}(f)); (3) We then deform the reference model to fit the CAESAR model. The vertices template and the CAESAR model are brought to one-to-one correspondence using an unsupervised non-rigid point set registration algorithm. The goal of 3D point set registration is to establish correspondence and recover the transformation between vertices of the template mesh and the CAESAR model. Our 3D registration process is based on iterative gradient-based optimization of the energy function with three data cost terms: (i) cost of fitting the non-landmark vertices to the nearest surface point of the laser scan ($E_V$); (ii) cost of fitting of the landmark points ($E_L$); and (iii) the internal regularization term to preserve the shape ($E_R$). The combined cost function is given by:
\begin{equation}
E = \alpha E_V + \beta E_L + \gamma E_R 
\end{equation} 					
The weights $\alpha,\beta,\gamma$ are adjusted to balance the smoothness of the registered shape and accuracy of alignment. The optimization process is illustrated in \ref{fig:overview}(f)). Using this method, we have registered about 1500 CAESAR North American Standing scan data images. With this registration, the CAESAR scan data are now transformed to a common parametrization scheme.

\begin{figure}[t]
\begin{center}
\includegraphics[width= 0.98\linewidth]{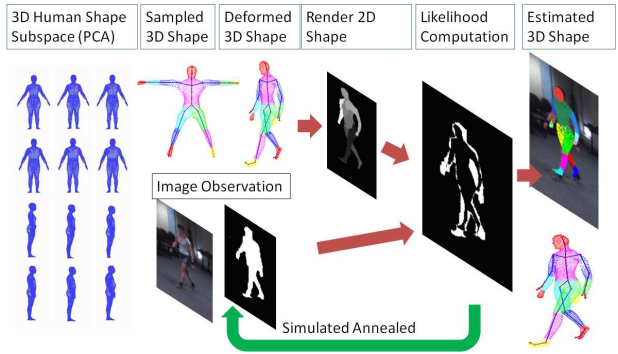} \\
\end{center}
\caption{Overview of shape estimation algorithm} 
\label{fig:ShapeEstimationOverview}
\end{figure}

For a test image sequence, we estimate the 3D shape of the human target by searching in the low-dimensional shape space learned using Principal Component Analysis (PCA). The 3D shape fitting algorithm essentially searches in the learned subspace of human 3D shapes for estimating best fitting PCA coefficients that has highest likelihood (same as used for pose optimization). Sampling the shape space however models anthropometric variability and can generate shapes of humans standing in a canonical pose. The shape is therefore non-rigidly deformed under the current pose for each sampled shape hypothesis. For doing the smooth deformation, each of the vertices in the 3D mesh is associated to multiple joints (less than a maximum of 6 joints). For optimizing the 3D shape, we use Annealed Particle Filtering to obtain optimal shape parameters best aligns with observation when projected to 2D image plane. \ref{fig:ShapeEstimationOverview} shows the entire 3D shape estimation framework. Anthropometric skeleton is critical to the accuracy of 3D shape fitting algorithm as it determines the alignment and realistic deformation of the 3D shape under the influence of skeletal pose. We estimate the skeleton for a 3D shape by estimating skeletal link lengths from the vertices and fitting the skeleton to the joint original locations using Levenberg-Marquardt(LM) optimizer. This optimization re-estimates the joint angles specific to the new skeleton and shape. 

\begin{figure}[t]
\begin{center}
\includegraphics[width= 0.98\linewidth]{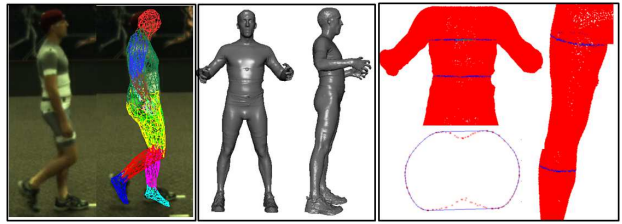}
\end{center}
\caption{Data used for evaluating 3D shape estimation framework. Part circumferences computed from the estimated 3D shapes are compared against groundtruth measurements estimated from laser scan shape of the same subject} 
\label{fig:PartDimensionEstimation}
\end{figure}

\noindent\textbf{Extracting BioSignatures:} Biosignatures extracted by our system include height, weight, gender and anthropometric measurements of the 3D human shape. Standard anthropometric distance measurements is done using geodesic distances. However, geodesic measurements are often difficult to simulate. For instance, the CAESAR neck base circumference is determined by resting an adjustable chain necklace on the subject, then measuring the length of the chain. Such a procedure would require a full physical simulation to achieve in software.  Hence, we restrict ourselves to a more tractable class of geodesic measurements: horizontal body part circumferences, in particular, the chest, waist, hips, right thigh, and right calf, as shown in \fig{fig:PartDimensionEstimation}(right)). On a 3D scan, these measurements can be approximated by finding a curve of intersection between the mesh and the plane of measurement(plane parallel to ground plane), finding the 2D convex hull to better simulate the taut tape or band, and measuring the length of the closed curve. The first step, finding the intersection of the plane and mesh, requires some filtering to ensure only the correct body part was measured. For the groundtruth laser scan data, we manually assign the which vertices correspond to which body part. In addition, we also manually assign part labels to each vertex for an average shape (one time labeling). Notice in \fig{fig:ShapeEstimationOverview} different parts of the human body are color coded. This allow the intersection to cover vertices of a specific body part. Additionally, limit marker vertices were chosen for each measurement, denoting the vertical extents of the region to be measured.  This step ensured that the chest would be measured below the armpit, the thigh below the crotch, etc. The results of these filtering steps for the chest are shown on the right in \fig{fig:PartDimensionEstimation}. For cases where the circumference was to be maximized or minimized, the smoothness of the function was exploited by using Levenberg-Marquardt optimization to quickly find the optimal height at which the circumference is maximum. Finally, taking the convex hull of the initial intersection curve proved very important for accuracy; the hips in particular often have deep concavities in the regions of the buttocks and crotch, as shown in the cross-sectional view in \fig{fig:PartDimensionEstimation}.

\noindent\textbf{Height, Weight and Gender Estimation:} Height of a human body can be computed directly from the estimated 3D shape using specific vertices at the top (head) and bottom (feet) of the 3D mesh. However in most poses, the human shape appears bent or not perfectly aligned in a standing pose. Distances between the two vertices usually give a poor estimate of the height. Similarly, for computing weight of a human body, we can compute volume of each of the body part and use average mass density of different parts to estimate the weight. However, different subjects may have different part density and this may give inaccurate estimate of the weight of the subject. Therefore we employ learning based methods to estimate the height using overall 3D shape of the subject. Overall anthropometric shape of the human subjects is strongly correlated to its height, weight and gender. We use non-linear regression (Relevance Vector Machine) functions to classify the gender and predictor height and age from the shape coefficients. 
\section{Experiments}
\label{sec:experiments}
We evaluate our tracking algorithms on HumanEva data set and provide pose estimation accuracies in terms of joint angles and joint center locations. The data set contained 3 subject ($S_1$, $S_2$ and $S_3$) performing three different activities (Walking, Jogging and Boxing). We only used $C_2$ sensor for training and testing our system. One of the testing sequences also include data captured from $C_3$ sensor (a viewpoint not used in our training data). For error reporting and testing, we partitioned the data set into training and testing sets. From each activity sequence, the first 200 frames were used for testing and the rest was used for training the bottom-up models as well as the optimal proposal density $p(\xx_n | \xx_{n-1},\rr_n)$. Both the distributions were trained using optimal set of parameters selected using cross-validation with the validation set containing randomly selected $10\%$ of the training data.
  
\noindent\textbf{Feature Extraction:} We use only shape descriptors extracted from silhouettes to train the predictive models. Our initial experiments using silhouette shapes along with internal edges in the image descriptors as well as for likelihood computation gave worse results than using silhouette alone. In all the experiments, the results were generated using shape context histogram (SCH) as the input image descriptor for learning the predictive models. SCH is computed from outer contour by uniformly sampling 100 pixels from it and voting for 5 radial and 12 angular bins with relative distance of pixels ranging from $1/8$ to $3$ on a log scale. The shape context is a robust shape descriptor that encodes relative locations of the sampled pixels w.r.t. a reference pixel. The features are vector quantized by clustering them into $K=400$ prototype cluster centers and modeling the distribution of these features using normalized inverse distances from these learned prototypes .  

\noindent\textbf{HumanEva 3D Pose Representation:}  HumanEva data set represents an articulated 3D human body pose as set of 20 joint locations. Joint location data cannot be readily transferred to animate a deformable mesh. We there pre-process the data by fitting a skeleton with 30 joints ($\approx 55$ degrees of freedom) to extract Euler angles for each joint. The skeleton for each dataset was estimated as average link lengths over first 100 frames of the motion capture data. We used these joint angles as groundtruth for validation of our framework. The average loss of joint location accuracy due to skeleton fitting ranged from $5$-$7$mm. The skeleton is fitted using the LM based damped least square  optimization. In doing so, we impose angular limit constraints to accurately estimate feet and wrist joints (not present in the HumanEva dataset). These are useful for overcoming ambiguity in twist angles of some of the joints. The global orientation of the human body is represented in cyclic co-ordinates using $cos/sin$ transform. 3D pose data in the original joint angle space has high dimensionality ($\approx$ 90) and is reduced to 5 dimensions using SLVM\cite{KSM07b}. Separate SLVM is trained for each activity and provides bi-directional mapping between the ambient and the latent space.  The overall parameter space of 3D pose is 11 dimensional (6 due to rigid body motion and 5 due to 3D pose).

\noindent\textbf{Predictive Models:} Both the predictive distributions $p_B(\xx | \rr)$ and $p(\xx_{n} | \xx_{n-1}, \rr_{n})$ are learned using Bayesian Mixture of Experts. $p_B(\xx | \rr)$ is modeled using two-level hierarchical Bayesian Mixture of Expert (hBME)  model. At the first level, we cluster the data points based on global pose orientation of the human target and train a classifier to recognize the orientation of the human body with respect to camera image plane. We quantize the $360^o$ human orientation span into 8 views and train a classifier to recognize the view based on the shape descriptor. At the second level, we train 2 view dependent expert predictors to output 3D pose.  $p(\xx_{n} | \xx_{n-1}, \rr_{n})$ is however trained using a simpler Bayesian Mixture of Expert model with 5 experts. As the predictors output the points in the decorrelated latent state space, we treat independent hBME for each latent space dimension. Both for learning classification and regression models, we used Relevance Vector Machine\cite{T01}. For each viewpoint, we also learn 3 view dependent regression functions to estimate exact orientation of the human. 

\noindent\textbf{Likelihood Computation and Hardware Optimization:} Likelihood computation is the costliest operation in the generative tracking. For the likelihood function during tracking, we use an average 3D human shape and deform it using an averaged sized skeleton. Using average shape regularizes the 3D pose optimizer at each time step and overcome local optima due to specialized 3D shape. A one-time manual skeletal alignment and weight-painting process is required for generating arbitrary human shapes and poses. We employed 200 particles in the PF tracker, with 10 simulated annealing iterations at each time step. As particle filtering involves independent computation of the likelihood function of the $N$ particles, we can parallelize the processing. We use efficient Graphics Processing Unit (GPU) based implementation for computing the likelihood function which is the bottleneck operation for the entire optimization process.

\begin{figure}[t]
\begin{center}
\includegraphics[width= 0.98\linewidth]{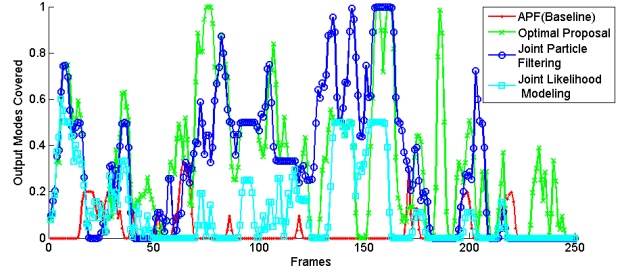}
\end{center}
\caption{Multimodal distribution propagation, ratio of degree of multimodality propagated by each of the particle filtering algorithm\textbf{\it(Best viewed in color)}} 
\label{fig:MultimodalDistribution}
\end{figure}

\noindent\textbf{Sustaining Multimodality in PF:} To characterize and compare the degree of modality propagated by different particle filtering algorithms, we use the weighting scheme in eqn. \ref{eq:multimodality} to weigh a correspondence between input and output cluster. For the input cluster corresponding to the input data, we compute which of the associated output clusters have been observed (or covered) for the current set of particles. For each association between the input and output cluster, we add the corresponding weight and compute the ratio with the maximum weight. Fig. \ref{fig:MultimodalDistribution} shows the degree of multimodality preserved by different PF algorithms for the jogging sequence. Note that baseline APF has minimum modality preserved compared to the other three PF enhancements. 

\begin{figure}[h]
\begin{center}
\includegraphics[width=0.98\linewidth]{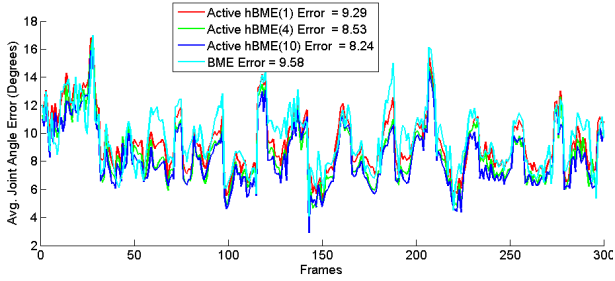} 
\end{center}
\caption{Online active learning, pose prediction accuracy of online active learning of the predictive models \textbf{\it(Best viewed in color)}} 
\label{fig:ActiveLearning}
\end{figure}

\noindent\textbf{Online Active Learning of Predictive Models:} For adapting the hBME model to the test domain, we apply Bayesian relevance based updating to the two levels of gate distributions and the expert regression functions. From a set of particle hypotheses, we select $N = 1-5\%$ with highest likelihood weights to update the bases sets of the gates $g_v$, $g_e$ and the Experts. For $N > 5\%$ approximate poses caused degradation in the accuracy of the predictors. Gate cluster for  $g_v$ and $g_e$ are identified based on viewpoint and proximity to expert predictors. After the parameter and bases updation, EM-iterations are run to refine the gates and expert cluster distributions. The entire updation mechanism is fast and performed in online fashion. Fig. \ref{fig:ActiveLearning} shows the average joint angle prediction accuracy in degrees for the hBME model for a jogging sequence on subject $S_3$. We used JPF for tracking and updated hBME at every frame using 1, 4 and 10 particles with highest weights. 
Notice also that the average error for each of the active learning scheme decreases using more number of particles to update the hBME model.    

\noindent\textbf{Pose Estimation Results:} Table \ref{tab:HumanEvaResults} shows quantitative evaluation results of our framework on the HumanEva dataset for the sensors $C_2$ and $C_3$. Since we trained only on data corresponding to sensor C2, for evaluating pose estimation results in sensor $C_3$, we estimated 3D pose using calibration parameters of sensor $C_2$. To compute errors, we transform the estimated root joint orientation by first removing the camera rotation due to $C_2$ and applying the rotation due to $C_3$. That is for the camera projection matrices: $\PP_{C_2} = [\RR_{C_2} \TT_{C_2}]$ and  $\PP_{C_3} = [\RR_{C_3} \TT_{C_3}]$, the pose is transformed using the rotation matrix $\RR_{C_3}\dot\RR^T_{C_2}$ before computing the error scores.
\begin{table} 
\label{tab:HumanEvaResults}
{\small 
{
\begin{tabular}{|c|c|c|c|c|}
\hline
Algorithm  Seq.  		&  {\bf APF}   & {\bf OPF}   & {\bf JPF}      & {\bf JLM}   \\
\hline\hline
Jog($S_1$ in $C_2$) (Joint Loc.)  	& 38.48  & 78.39  & {\bf 34.55} & 41.24  \\
Jog($S_1$ in $C_2$) (Joint Angle) 	& 9.32   & 12.54  & {\bf 8.19} & 12.11  \\
\hline
Jog ($S_2$ in $C_2$) (Joint Loc.)  	& 43.04 & 35.05  & {\bf 31.01} & 58.58   \\
Jog ($S_2$ in $C_2$) (Joint Angle) 	& 11.07 & 9.50  & {\bf 7.25} & 9.06  \\
\hline
Jog ($S_3$ in $C_2$)(Joint Loc.)  	& 78.18  & 75.41  & {\bf 38.74} & 62.57 \\
Jog ($S_3$ in $C_2$)(Joint Angle) 	& 11.03  & 11.26 & {\bf 9.06} & 11.55\\
\hline\hline
Box ($S_2$ in $C_2$)(Joint Loc.)   	& 67.4 & 34.73  & 43.58 & {\bf 27.65}  \\
Box ($S_2$ in $C_2$)(Joint Angle)  	& 18.18 & 12.55  & 14.08 & {\bf 8.39}  \\ 
\hline
Box ($S_1$ in $C_2$)(Joint Loc.)   	& 43.56  & 33.27 & 25.19 & {\bf 23.12} \\
Box ($S_1$ in $C_2$)(Joint Angle)  	& 13.22  & 11.70  & 10.05  & {\bf 7.05} \\ 
\hline
Box ($S_3$ in $C_2$)(Joint Loc.)   	& 49.61   & 68.6 &  {\bf 37.37} & 55.15 \\
Box ($S_3$ in $C_2$)(Joint Angle)  	& 15.41  & 23.75 & {\bf 12.11} & 13.02 \\ 
\hline\hline
Walk ($S_1$ in $C_2$)(Joint Loc.)   	& 26.43 & 30.42 & {\bf 25.01} & 26.64 \\
Walk ($S_1$ in $C_2$)(Joint Angle)  	& 7.61  & 7.23  & 5.04  & {\bf 4.10} \\ 
\hline
Walk ($S_2$ in $C_2$)(Joint Loc.)   	& 60.40  & 37.04 &  {\bf 34.61 }& 35.06 \\
Walk ($S_2$ in $C_2$)(Joint Angle)  	& 9.71  & 9.06  & {\bf 6.01}  & 6.74\\ 
\hline
Walk ($S_3$ in $C_2$)(Joint Loc.)   	& 54.09  & 63.09 & {\bf 27.61} & 64.25\\
Walk ($S_3$ in $C_2$)(Joint Angle)  	& 9.52  & 9.32  & {\bf 4.60} & 7.49 \\ 
\hline
Jog ($S_2$ in $C_3$) (Joint Loc.)  	& 51.43 & 40.12  & {\bf 38.91} & 54.33 \\
Jog ($S_2$ in $C_3$) (Joint Angle)  	& 11.49 & 12.57 & {\bf 10.5}  & 13.28  \\ 
\hline
\end{tabular}
}}
\caption{3D pose estimation accuracies in average joint location error and joint angle error, for various PF algorithms. Highlighted values denote the best of the 4 algorithms that include: APF - Annealed Particle Filtering,  learned Optimal Proposal Density based PF, JPF - Joint Particle Filtering and JLM - Joint Likelihood Modeling.  JPF clearly outperforms the baseline APF and the other two improvements proposed in the work}
\label{tab:algorithm}
\end{table}
The results in Table \ref{tab:HumanEvaResults} demonstrates the improved accuracies both in terms of joint location and joint angles. Some significant discrepancies between the error rates of joint locations and joint angles  is because joint angle error does not take into account the error in the orientation of the pose but only the body joint angles. Even a small error in the root joint can cause sizable difference in the joint location but none in the angles of the joints.  Based on the results, JPF clearly outperforms both the baseline algorithm based on APF and in most cases PF based on JLM and Optimal Proposal density. Higher accuracy of JPF is due to very detailed bottom-up predictive models (total  $5$ Latent Dim. $\times 8$ Views $\times 2$ Experts  regression functions in hBME and $8$ Views $\times 3$ Experts regression functions for orientations). The bottom-up proposal density with 16 Gaussian components can efficiently represent any depth and view based ambiguity to provide diverse set of samples having higher weights and closer to the true posterior. APF based on learned optimal proposal density performs well on certain sequences, however for other sequences it may output states far from the training data, causing it to recursively output mean predictions (as the combined feature and state prediction from previous step significantly differs from the training exemplars). The errors are usually difficult to recover from.  JLM based APF in most cases outperforms baseline APF and Optimal Proposal Density based APF. \Fig{fig:CompareAPF} compares the pose estimation results from the four trackers. Notice that JPF is able to overcome the errors due to view-based and left-right leg forward ambiguities. 
The generic bottom-up model can be be applied to estimate pose in any orientation. 

\begin{figure}[t]
\begin{center}
\includegraphics[width= 0.98\linewidth]{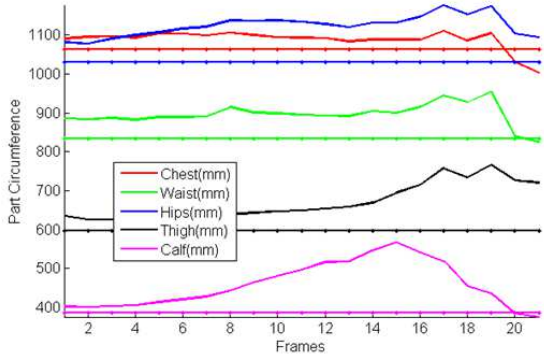}
\end{center}
\caption{Shape estimation results for a subject performing walking motion. We compare the error in the estimated circumference of various body parts (Chest, Waist, Hips, Thigh and Calf) with respect to groundtruth circumferences computed from the laser scan data of the subject} 
\label{fig:PartDimensionEstimationResult}
\end{figure}

\noindent\textbf{Shape Estimation Results:} In order to evaluate the accuracy of our 3D shape estimation framework, we use laser scan data of a subject as the groundtruth shape and apply our shape estimation algorithm to reconstruct its 3D shape for a walking motion image sequence (see  \fig{fig:PartDimensionEstimation}). We use the 3D body part measurements of the laser scan data to compute the error in 3D body part circumference estimation. \Fig{fig:PartDimensionEstimationResult} shows the plot of circumferences estimated from the fitted 3D body shape for some frames of the image sequence, and corresponding groundtruth measurements. Table \ref{tab:partCircumferences} shows the comparison of the groundtruth body part girths and the circumferences computed from the estimated 3D body shape and averaged over 20 frames.
\begin{table} 
{\small 
{
\begin{tabular}{|c|c|c|c|c|c|}
\hline
Measurement  &  {\bf Chest}  & {\bf Waist} & {\bf Hips}& {\bf Thigh} & {\bf Calf}  \\
    	     &  {\bf (cm)}   & {\bf(cm)}   & {\bf(cm)} & {\bf (cm)}  & {\bf (cm)} \\
\hline
Groundtruth   	&  106.4     &  83.46   &  103.18  & 59.6      & 38.65\\
Estimated       &  108.85    &  89.59   &  112.44  & 67.08     & 45.76 \\
Error(\%)   	&  2.3\%     &  7.35\%  &   8.98\% &  12.55\%  & 18.41\%\\
\hline
\end{tabular}
}}
\caption{Part circumference estimation accuracy}
\label{tab:partCircumferences}
\end{table}

\begin{figure}[h]
\begin{center}
\includegraphics[width= 0.98\linewidth]{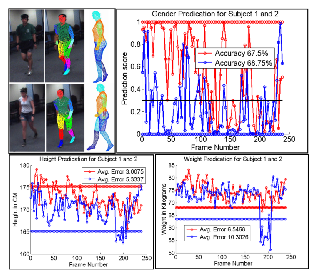}
\end{center}
\caption{Attributes estimation results for subjects 1 and 2} 
\label{fig:AttributesEstimation1}
\end{figure}

\begin{figure}[h]
\begin{center}
\includegraphics[width= 0.98\linewidth]{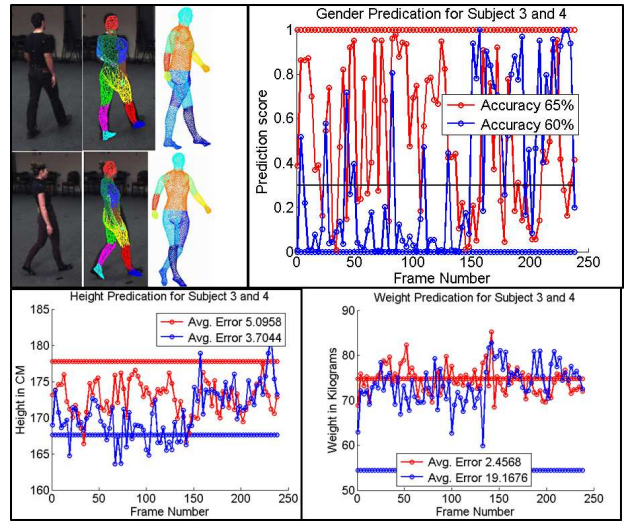}
\end{center}
\caption{Attributes estimation results for subjects 3 and 4} 
\label{fig:AttributesEstimation2}
\end{figure}

\noindent\textbf{Attributes Estimation:} Attribute estimation accuracy was evaluated on 4 targets. 3D shapes fitted to 250 frames of the video sequence were used to infer attributes using the learned regression functions. \Fig{fig:AttributesEstimation1} shows the plots of the results on the first two subjects where subject 1 is male and subject 2 is female. \Fig{fig:AttributesEstimation2} shows the same for the subjects 3 and 4, where subject 3 is a male and subject 4 is female. For gender prediction the classifier gave the score of being a male, that gave best accuracy when the threshold is set $0.3$. Table \ref{tab:AttribEstimationResults} shows the average prediction error for height and weight, and prediction accuracy for gender for the 250 frames. We also extracted 20 frames from each of the subject sequence that had best shape fitting likelihood. The average prediction accuracy significantly improved when only best fitted shapes were used for attributes estimation as shown in the results in parentheses in table \ref{tab:AttribEstimationResults}.             
\begin{table} 
{\small
{
\begin{tabular}{|c|c|c|c|}
\hline
Error/      	&  {\bf Height} & {\bf Weight}   & {\bf Gender}  \\
 Subject    	&  {\bf Error(cm)}   & {\bf Error (Kgs) }   & {\bf Accuracy(\%)} \\
\hline
Subject 1   	&  3.0(2.1)     &  6.55 (4.32)  &  67.5\%(72.1\%) \\
Subject 2   	&  5.33(3.74)   &  10.3 (9.9)  &  68.75\%(77.5\%) \\
Subject 3   	&  5.10(3.22)   &  2.46 (2.9)  &  65.0\% (70.5\%)\\
Subject 4   	&  3.70(2.23)   &  19.17(15.3)  &  60.3\%(67.9\%)\\
\hline
\end{tabular}
}}
\caption{Attributes prediction accuracy}
\label{tab:AttribEstimationResults}
\end{table}

\section{Conclusion}
We have developed a fully automated system for 3D pose and shape estimation and analysis from monocular image sequences. We develop three principled approaches to enhance particle filtering  by integrating bottom-up information  either as proposal density for obtaining more diverse particles or as complementary cues to improve likelihood computation during the correction step. Through extensive experimental evaluation we have demonstrated that our algorithms enhance the ability of the particle filtering to propagate multimodality for effective reconstruction of 3D poses from 2D images. In this work, we also demonstrated that a feedback mechanism from top-down modeling can further adapt and improve the bottom-up predictors to enhance the overall tracking performance.

\noindent\textbf{Acknowledgements:} This work was supported by Air Force Research Lab, contract number FA8650-10-C-6125.
  
{
\bibliographystyle{abbrv}
\bibliography{papbib}  

\begin{thebibliography}{10}

\bibitem{AMGC02}
S.~Arulampalam, S.~Maskell, N.~Gordon, and C.~T.
\newblock A tutorial on particle filters for online nonlinear/non-gaussian
  bayesian tracking.
\newblock {\em Signal Processing, IEEE Transactions on}, 2002.

\bibitem{CAESAR}
Dataset.
\newblock Caesar: Civilian american and european surface anthropometry resource
  project.
\newblock In {\em http://store.sae.org/caesar/}, volume~1, 2002.

\bibitem{DBR00}
J.~Deutscher, A.~Blake, and I.~D. Reid.
\newblock Articulated body motion capture by annealed particle filtering.
\newblock In {\em CVPR}, 2000.

\bibitem{Doucet00}
A.~Doucet, S.~Godsill, and C.~Andrieu.
\newblock On sequential monte carlo sampling methods for bayesian filtering.
\newblock {\em STATISTICS AND COMPUTING}, 2000.

\bibitem{Doucet11}
A.~Doucet and A.~M. Johansen.
\newblock A tutorial on particle filtering and smoothing: fifteen years later,
  2011.

\bibitem{IB98a}
M.~Isard and A.~Blake.
\newblock Icondensation: Unifying low-level and high-level tracking in a
  stochastic framework.
\newblock In {\em ECCV}, 1998.

\bibitem{KSM07b}
A.~Kanaujia, C.~Sminchisescu, and D.~Metaxas.
\newblock Spectral latent variable models for perceptual inference.
\newblock {\em ICCV}, 2007.

\bibitem{LC04}
M.~W. Lee and I.~Cohen.
\newblock Proposal maps driven mcmc for estimating human body pose in static
  images.
\newblock In {\em CVPR (2)}, pages 334--341, 2004.

\bibitem{LN09}
M.~W. Lee and R.~Nevatia.
\newblock Human pose tracking in monocular sequence using multilevel structured
  models.
\newblock {\em Trans. Pattern Anal. Mach. Intell.}, 2009.

\bibitem{RS01}
R.~Rosales and S.~Sclaroff.
\newblock Learning body pose via specialized maps.
\newblock In {\em NIPS}, pages 1263--1270, 2001.

\bibitem{SU10}
M.~Salzmann and R.~Urtasun.
\newblock Combining discriminative and generative methods for 3d deformable
  surface and articulated pose reconstruction.
\newblock In {\em CVPR 2010}. IEEE, 2010.

\bibitem{SBB07}
L.~Sigal, A.~O. Balan, and M.~J. Black.
\newblock Combined discriminative and generative articulated pose and non-rigid
  shape estimation.
\newblock In {\em NIPS}, 2007.

\bibitem{SHSV10}
L.~Sigal, A.~O. Balan, and M.~J. Black.
\newblock Humaneva: Synchronized video and motion capture dataset and baseline
  algorithm for evaluation of articulated human motion.
\newblock {\em Technical Report}, 2010.

\bibitem{SKLM05}
C.~Sminchisescu, A.~Kanaujia, Z.~Li, and D.~N. Metaxas.
\newblock Discriminative density propagation for 3d human motion estimation.
\newblock In {\em CVPR (1)}, 2005.

\bibitem{SKM06}
C.~Sminchisescu, A.~Kanaujia, and D.~N. Metaxas.
\newblock Learning joint top-down and bottom-up processes for 3d visual
  inference.
\newblock In {\em CVPR (2)}, 2006.

\bibitem{T01}
M.~E. Tipping.
\newblock Sparse bayesian learning and the relevance vector machine.
\newblock {\em Journal of Machine Learning Research}, 1, 2001.

\bibitem{VDP03}
J.~Vermaak, A.~Doucet, and P.~Perez.
\newblock Maintaining multimodality through mixture tracking.
\newblock In {\em ICCV}, 2003.

\end{thebibliography}
%
%
}
\end{document}